\documentclass[journal]{IEEEtran}

\usepackage{mathtools}
\usepackage{subfigure}
\usepackage{amsmath}
\usepackage{graphicx}
\usepackage{multirow}
\usepackage{makecell}
\usepackage{float}
\usepackage{cite}
\usepackage{indentfirst}
\usepackage{amsthm}
\usepackage{color}
\usepackage{mathrsfs}
\usepackage{amssymb}
\usepackage{amsbsy}
\usepackage{times}  
\usepackage{url}  
\usepackage{verbatim}
\usepackage{hyperref}
\usepackage{hyperref}
\usepackage{algorithm}
\usepackage{algorithmic}

\newtheorem*{theorem*}{Theorem}

\newtheorem*{corollary*}{Corollary}

\newtheorem*{lemma*}{Lemma}

\newtheorem*{proposition*}{Proposition}

\ifCLASSINFOpdf

\else

\fi

\begin{document}

\title{ Coordinated Transformer with Position \& Sample-aware Central Loss for Anatomical Landmark Detection}

\author{Qikui Zhu, Yihui Bi, Danxin Wang,  Xiangpeng Chu, Jie Chen, Yanqing Wang
\thanks{The source code of this work is available at the \href{https://github.com/ahukui/CoorTransformer}{GitHub repository.}}%
\thanks{Q.~Zhu is with Department of Biomedical Engineering, Case Western Reserve University, OH, USA. (e-mail: QikuiZhu@163.com).}%
\thanks{Y.~Bi is with Department of Orthopaedics, The Second Affiliated Hospital of Anhui Medical University, Hefei 230601, China. Institute of Orthopaedics, Research Center for Translational Medicine, The Second Affiliated Hospital of Anhui Medical University, Hefei 230601, China (e-mail: docbyh1992@163.com).}%
\thanks{D.~Wang is with the College of Computer Science and Technology, China University of Petroleum, Qingdao, China (e-mail: wangdx@upc.edu.cn).}%
\thanks{X.~Chu is with Guangzhou Twelfth people’s Hospital, Guangzhou Occupational Disease Prevention and Treatment Hospital, Guangzhou Otolaryngology-head and Neck Surgery Hospital, Guangzhou, China. (e-mail: 13155182545@163.com).}%
\thanks{J.~Chen is with the Department of Pathology and Institute of Clinical Pathology, West China Hospital, Chengdu, China.(e-mail: jzcjedu@foxmail.com).}
\thanks{Y.~Wang is with Department of Gynecology, Renmin Hospital of Wuhan University, Wuhan, China. (e-mail: yanqingwang543@gmail.com).}%

}

\maketitle

\makeatletter
\def\ps@IEEEtitlepagestyle{
	\def\@oddfoot{\mycopyrightnotice}
	\def\@evenfoot{}
}
\def\mycopyrightnotice{
	{\footnotesize
		\begin{minipage}{\textwidth}
			\centering
			Copyright~\copyright~2019 IEEE. Personal use of this material is permitted.  Permission from IEEE must be obtained for all other uses, in any current or future media, including reprinting/republishing this material for advertising or promotional purposes, creating new collective works, for resale or redistribution to servers or lists, or reuse of any copyrighted component of this work in other works.
		\end{minipage}
	}
}

\begin{abstract}
Heatmap-based anatomical landmark detection is still facing two unresolved challenges: 1) inability to accurately evaluate the distribution of heatmap;
2) inability to effectively exploit global spatial structure information.
To address the computational inability challenge, we propose a novel position-aware and sample-aware central loss. Specifically, our central loss can absorb position information, enabling accurate evaluation of the heatmap distribution. More advanced is that our central loss is sample-aware, which can adaptively distinguish easy and hard samples and make the model more focused on hard samples while solving the challenge of extreme imbalance between landmarks and non-landmarks.
To address the challenge of ignoring structure information, a Coordinated Transformer, called CoorTransformer, is proposed, which establishes long-range dependencies under the guidance of landmark coordination information, making the attention more focused on the sparse landmarks while taking advantage of global spatial structure. Furthermore, CoorTransformer can speed up convergence, effectively avoiding the defect that Transformers have difficulty converging in sparse representation learning.
Using the advanced CoorTransformer and central loss,
we propose a generalized detection model that can handle various scenarios, inherently exploiting the underlying relationship between landmarks and incorporating rich structural knowledge around the target landmarks.
We analyzed and evaluated CoorTransformer and central loss on three challenging landmark detection tasks. The experimental results show that our CoorTransformer outperforms state-of-the-art methods, and the central loss significantly improves the performance of the model with $p$-values $< 0.05$. 
\end{abstract}


\IEEEpeerreviewmaketitle

\begin{figure*}[!t]
  \centering
\includegraphics[width=0.75\textwidth]{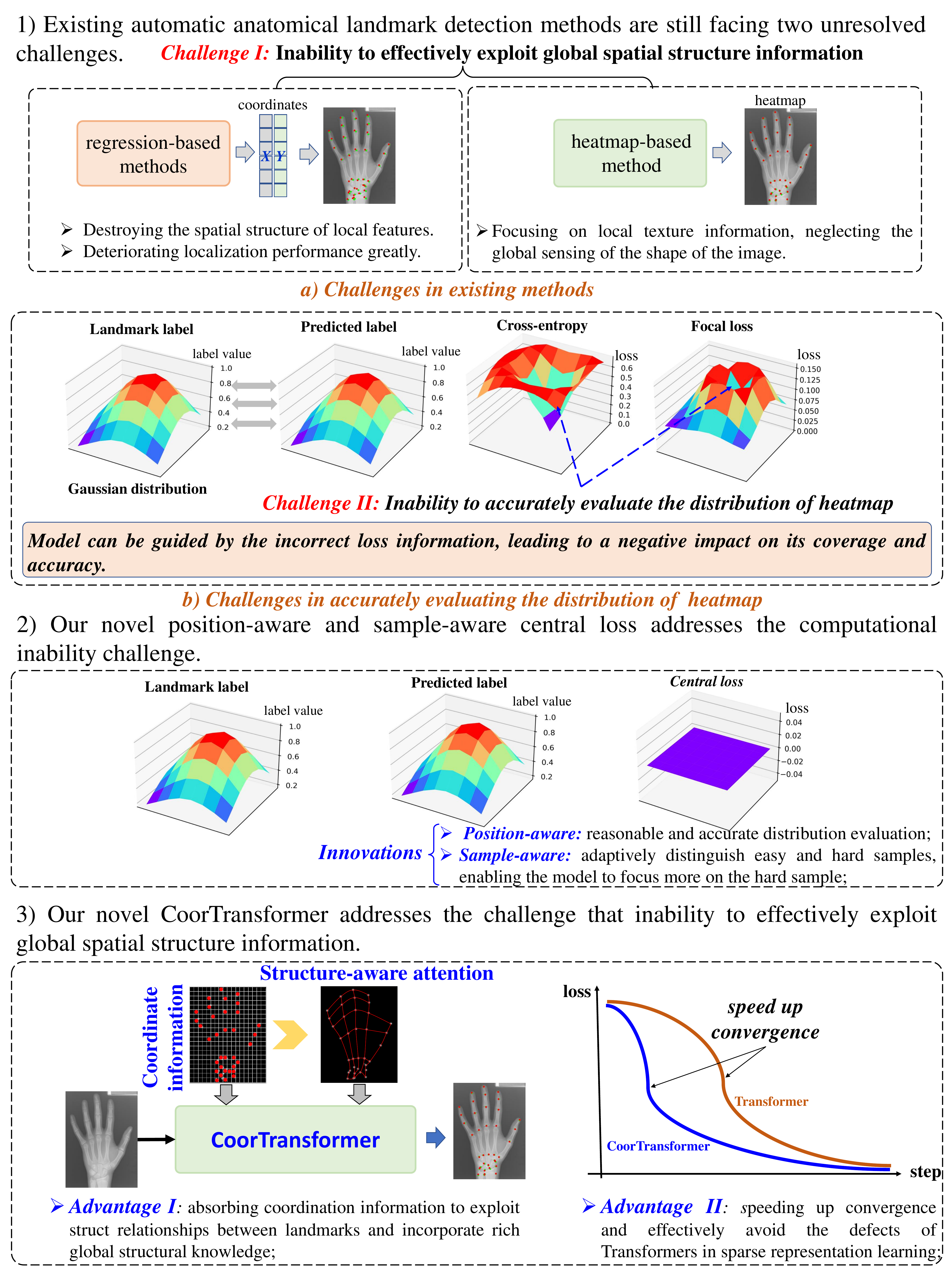}
  \caption{1) Existing automatic anatomical landmark detection methods are still facing two unresolved challenges: a) inability to effectively
exploit global spatial structure information; b)  inability to accurately
evaluate the distribution of heatmap. 2) Our novel central loss can address the computational inability challenge by a position-aware mechanism and a sample-aware mechanism, which can absorb position
information for accurate evaluating distribution and adaptively distinguish easy and
hard samples, enabling the model to focus more on the hard samples. Additionally, 3) our CoorTransformer overcomes the challenge of  inability to effectively
exploit global spatial structure information by novelty incorporating rich global structural knowledge.
}
  \label{example}
\end{figure*}
\section{Introduction}
\label{sec:intro}
\IEEEPARstart{A}{ccurate}  and automatic detection of anatomical landmarks plays an essential role in the field of biomedical research, as it has contributed to the development of various medical fields~\cite{rahmatullah2012image,gertych2007bone}. For example, to plan cardiac interventions, it is essential to identify standardized planes of the heart; correspondences between anatomical landmarks in medical images can offer supplementary guidance information for deformable medical image registration~\cite{grewal2020end}.
In clinical settings, landmarks are typically annotated manually. Nevertheless, manual annotation of landmarks is often a tedious and time-consuming task that requires significant expertise in anatomy and is susceptible to inter- and intra-observer errors. Hence, there is an urgent need for an accurate and automatic method for  anatomical landmarks detection.

Although existing approaches~\cite{ibragimov2014shape,vstern2016local,lindner2014robust,urschler2018integrating,zhu2021you}, including regression-based methods and heatmap-based methods, have significantly improved landmark detection, they tend to focus more on local texture information and disregard global spatial structure information (Challenge I), as depicted in Figure.~\ref{example}(a). Specifically, regression-based methods try to map the input image to the landmark coordination via fully connected layers. However, the fully connected layers would destroy the spatial structure of local features and further make the localization performance deteriorates greatly. Although heatmap-based models can preserve the local spatial structure of image features and achieve better performance than coordination regression-based models generally, they are more concerned with local texture information and neglect the global sensing of the shape of the image, particularly for the underlying relationship between landmarks, which makes them vulnerable to large appearance variations. Furthermore, another unsolved challenge in heatmap-based detection is inability to accurately evaluate the distribution of heatmap (Challenge II). As the label of landmarks has been transformed into probability distributions, existing loss functions, such as cross-entropy loss and focal loss~\cite{lin2017focal}, can not provide an accurate evaluation for each pixel (Figure.~\ref{example}(b)). Therefore, several issues in landmark detection require urgent attention and resolution.

To address the computational inability challenge for heatmap-based landmark detection, we propose a novel position-aware and sample-aware central loss (Figure.~\ref{example}(2)). Specifically, central loss addresses three challenges in heatmap-based landmark detection: 1) Inability to accurately evaluate the distribution of heatmap: As the landmark point is converted to a Gaussian probability, existing losses cannot accurately evaluate the distribution. 2) Inability to distinguish easy and hard samples: differences in location and surrounding tissue make landmarks specific, and it's necessary to distinguish between easy and hard landmarks.
3) Inability to handle extreme imbalance: in the landmark, there is an extreme imbalance between landmark pixels and background pixels. A large number of negatives can be overwhelming and make up the majority of the loss, resulting in the gradient being dominated by large negative values and ignoring the landmark information. Even when adding weight to balance the importance of positive/negative pixels, it does not differentiate between easy/hard pixels.
Furthermore, and most importantly, existing losses such as cross-entropy loss and focal loss are not perfectly suitable for scenarios where the ground truth is in the form of probability distributions. As shown in Figure~\ref{example}(b), these loss functions may produce incorrect loss information about the model's performance. This can result in the model being guided in the wrong direction, ultimately affecting its coverage and accuracy.
Our position-aware and sample-aware central loss solves the above challenges by using two novel terms: 1) self-weighting term; 2) difference-modulating term. The self-weighting term is associated with the relative position of each pixel. Pixels that are closer to the landmark are assigned greater weights, which makes the central loss is position-aware and advances the balancing of the influence of different pixels for addressing the challenge of extreme imbalance. The difference-modulating term formulates the distance between the predicted value and the ground truth, providing accurate distribution evaluation to guide the model toward the optimal convergence. This term brings two contributions: 1) it can accurate evaluate the labels with probability values, overcoming the computational inability challenge, and 2) it's sample-aware, which can distinguish easy/hard samples and pay more attention to the hard samples while solving the extreme imbalance challenge. (More details about the relationship between cross-entropy loss and focal loss in Section III.B.)

Considering inability to effectively exploit global spatial structure information, we propose a Coordinated Transformer (Figure.~\ref{example}(3)), called CoorTransformer, for spatial structure exploitation. Although Transformer's~\cite{vaswani2017attention} attention mechanism is capable of encoding long-range dependencies or heterogeneous interactions~\cite{zhu2022multi,valanarasu2021medical,hatamizadeh2022unetr}, Transformer is known to be effective only when large training datasets are available. What's more, as the attention module assigns nearly uniform attention weights to all pixels in the feature maps, long training epochs and enough training samples are required for the Transformer to focus on sparse landmarks. Due to the above-mentioned problems, existing Transformers are not efficient enough to extract meaningful information from the limited and sparse landmarks and meet the defect that difficult to converge. Superior to existing Transformers, CoorTransformer novelty uses coordination information when establishing long-range dependencies between landmarks. The coordination information makes the attention more concerned with spares landmarks, meanwhile, incorporating rich global structural knowledge.
What's more, under the guidance of the coordination information, CoorTransformer can quickly and accurately locate and identify sparse landmarks, which can speed up convergence and effectively avoid the defects of Transformers in sparse representation learning. Specifically, our CoorTransformer improves the ability of Transformer to extract the most relevant features from the feature maps in the vicinity of the target landmark and exploit the underlying relationship among landmarks for incorporating rich structure knowledge and quicker converge.



A generalized detection model that can handle various  scenarios is built upon the CoorTransformer and central loss. The details of the detector are described in section IV.B. To better evaluate the advancement of CoorTransformer and central loss, three challenging landmark detection tasks~\cite{wang2016benchmark,zhu2021you,jaeger2013automatic} are employed. Experimental results show that our central loss can improve the accuracy of landmark detection and our CoorTransformer-based detector outperforms the state-of-the-art methods.

Our contributions include:
\begin{itemize}
    \item For the first time, we address the issue of inability to accurately evaluate the distribution of heatmap by our novel designed position-aware and sample-aware central loss.

    \item Our novel central loss is both position-aware and sample-aware, overcoming computational limitations and enabling adaptive differentiation between easy and hard samples while solving the challenge of extreme imbalance.

    \item Our novel Coordinated Transformer makes significant advancements in exploiting the underlying relationship between sparse landmarks and incorporating rich structural knowledge by introducing coordination information. What's more, it avoids the defect of Transformer struggling to converge in sparse representation learning.

    \item A generalized detection mode is proposed, which overcomes the drawback of neglecting global sensing of spatial structure present in existing methods. Extensive experiments demonstrate that our CoorTransformer and central loss contribute to landmark detection under various medical images. It shows good generalization ability in the multi-dataset evaluation and achieves state-of-the-art accuracy on several landmark detection benchmarks.
\end{itemize}

\section{Related works}
\subsection{Landmark detection}
Accurate and automatic detection of anatomical landmark is an essential step for many image analysis and interpretation methods in the field of biomedical research. It has been widely used in various fields, including localizing tumors, detecting organs or structures, and searching for regions of interest in related medical images.

Recently, a number of automatic detection methods for anatomical landmarks have been introduced. For example,
Ibragimov et al.~\cite{ibragimov2014shape} proposed a novel landmark-based shape representation method, which is combined with the game-theoretic framework for landmark detection. Specifically, the connections among landmarks describing a 3-D object are established according to their cost to the resulting shape representation and by considering concepts from transportation theory.
Klinder et al.~\cite{KLINDER2009471} first reported an automatically whole-spine vertebral bone
identification, detection, segmentation method in CT images.
{\v{S}}tern et al.~\cite{vstern2016local} investigated different RRF architectures and proposed a novel random forest localization algorithm. The proposed algorithm can implicitly model the global configuration of multiple, potentially ambiguous landmarks and distinguish locally similar structures by automatically identifying and exploring the back-projection of the response from accurate local RF predictions.
Hanaoka et al.~\cite{hanaoka2017automatic} presented a novel Gibbs-sampling- and importance- sampling-based combinatory optimization framework for 197 landmarks detection in computed tomography (CT) volumes. Moreover, the proposed framework can  stochastically handle missing landmarks caused by detection failure.
L{\'o}pez-Linares et al.~\cite{lopez2018fully} presented a fully automatic approach for robust and reproducible thrombus region of interest detection and subsequent fine thrombus segmentation based on fully convolutional networks and holistically-nested edge detection network.
Alternatively, Ghesu et al.~\cite{ghesu2018towards} introduced a new paradigm for accurately detecting anatomical landmarks in 3D-CT scans with arbitrary field-of-view by using the concept of multi-scale DRL and robust statistical shape model. Nair et al.~\cite{nair2020exploring} utilized multiple uncertainty estimates based on Monte Carlo (MC) dropout and developed a 3D segmentation model for lesion detection and segmentation in medical images.
Vlontzos et al.~\cite{vlontzos2019multiple} formulated the problem of multiple anatomical landmark detection as a multi-agent reinforcement learning scenario and proposed a Collaborative DQN for landmark detection in brain and cardiac MRI
volumes and 3D US.
Chen et al.~\cite{chen2022semi} proposed a model-agnostic shape-regulated self-training framework for semi-supervised landmark detection by fully utilizing the global shape constraint. Specifically, a PCA-based shape
model is designed to adjust pseudo labels and eliminate abnormal ones for ensuring pseudo labels are reliable and consistent. And a novel Region Attention loss is proposed to make the network automatically focus on the structure-consistent regions around pseudo labels.
Lee et al.~\cite{lee2022cephalometric} proposed a single-passing convolutional neural network for accurate landmark detection. The proposed network first regresses the initial positions of all the landmarks for extracts of global contexts. Subsequently, extracting local features from landmark-centered patches through global regression. And a novel patch-wise attention module is proposed to weigh the relative importance of global and local features for better feature fusion.

\subsection{Transformer}
Recently,  Transformer~\cite{dosovitskiy2020image} emerges as an active research area in the computer vision community and has shown its potential to be a viable alternative to CNNs in many task. For example,
Li et al.~\cite{li2022towards} proposed a facial landmark detection network DTLD based on the cascaded transformer. The proposed method can directly regress landmark coordinates and can be trained end-to-end. Additionally, self-attention and deformable attention used in DTLD enable structure relationship exploring and more related image feature extracting.
Zhu et al.~\cite{zhu2022multi} introduced a multi-view coupled Transformer that helps overcome the limitation of the Transformer, where the dimensional relationship within the Transformer is not considered. Their approach includes two types of self-attention mechanisms that examine the relationship between spatial and dimensional attention, and a view-complementary approach that models both local and global spatial contextual information.
Zhu et al.~\cite{zhu2022datr} proposed a universal model for multi-domain landmark detection by taking advantage of Transformer for modeling long dependencies and develop a domain-adaptive transformer model
Wu et al.~\cite{wu2022transmarker} proposed a
pure Transformer based model for facial landmark detection.  Extensive experiments demonstrate that the proposed Transformer outperforms the most recent CNN-based
methods.
Wan et al.~\cite{wan2023precise}
utilized reference heatmap information and proposed a Reference Heatmap Transformer (RHT) for precise facial landmark detection. The proposed RHT consists of a Soft Transformation Module (STM), a Hard Transformation Module (HTM), and a Multi-Scale Feature Fusion Module (MSFFM). STM and HTM for extracting the reference heatmap information and facial shape constraints. MSFFM fuses the transformed heatmap features and the semantic features to enhance feature representations.
\begin{figure}[t]
  \centering
    \includegraphics[width=0.45\textwidth]{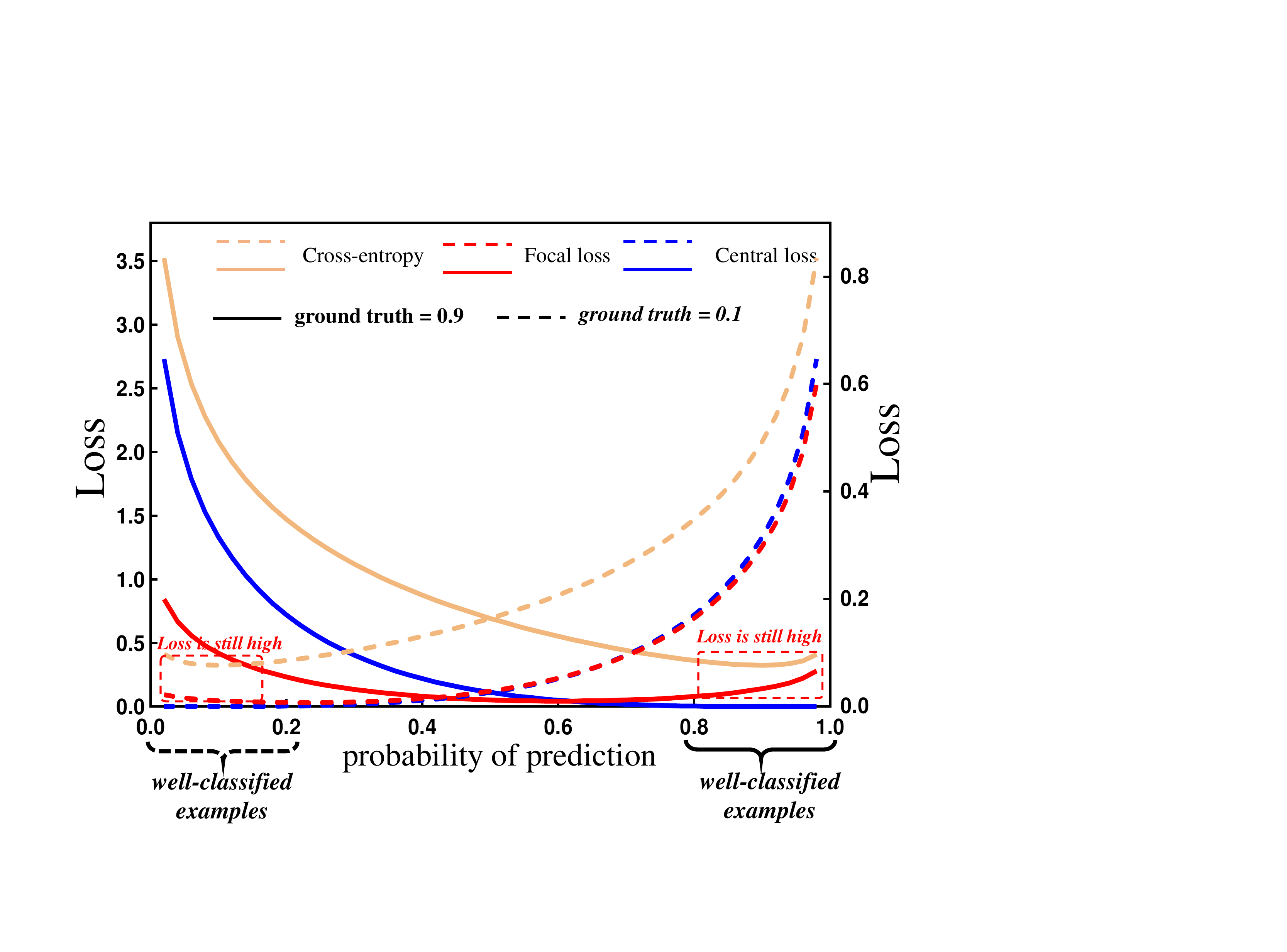}
  \caption{Comparison between cross-entropy, focal loss, and central loss functions, particularly when analyzing positive and negative samples with a distribution of 0.9 and 0.1, which demonstrates that our central loss function is  more computationally accuracy than cross-entropy and focal loss.}
  \label{fig_lossCM}
\end{figure}
\begin{figure*}[]
  \centering  \includegraphics[width=0.7\textwidth]{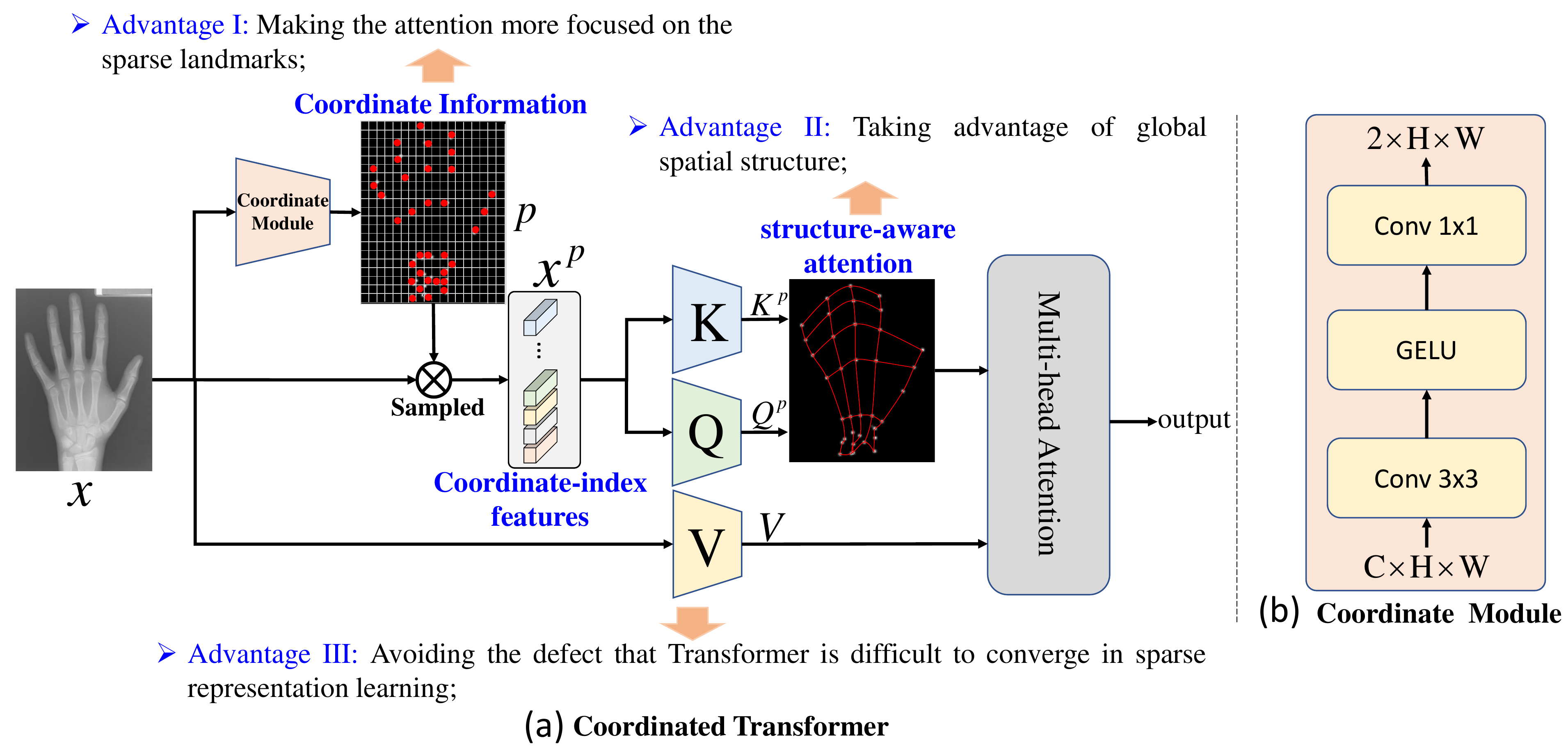}
  \caption{
Our CoorTransformer establishes long-range dependencies under the guidance of landmark coordination information, making the attention more focused on the sparse landmarks ({\bf Advantage I}) while taking advantage of global spatial structure ({\bf Advantage II}). Furthermore, CoorTransformer can speed up convergence, which effectively avoids the defect that Transformer is difficult to converge in sparse representation learning ({\bf Advantage III}). (a) A graphical illustration of Coordinated Transformer. (b) The detail of the coordinate module.
}
  \label{model}
\end{figure*}
\section{Central Loss}
\subsection{Definition of central loss}
The central loss is specifically designed to tackle three challenges in heatmap-based landmark detection tasks:  1)  inability to accurately evaluate the distribution of heatmap; 2) inability to distinguish easy/hard pixels; 3) difficulty dealing with extreme imbalance between landmark pixels and background pixels;

Formally, our central loss consists of two novel terms: (1) self-weighting term; (2) difference-modulating term, as described in Eq.~\ref{eq1}.
\begin{equation}
CL({p_x},{p_y}) =  - \underbrace {{p_y}}_{(1)} * \overbrace {{{\left| {{p_y} - {p_x}} \right|}^r}}^{(2)} * \log ({p_x})
\label{eq1}
\end{equation}
where, ${p_y} \in [0,1], {p_x} \in [0,1]$ is ground truth and predication results, respectively.
$r \ge 0$ is weighting parameter. Remarkable, the ground truth ${p_y}$ is a probability value between 0 and 1.


The self-weighting term is designed to balance the influence of positive and negative samples. Unlike the typical weighting factor, which is determined by inverse class frequency or set as a hyperparameter through cross-validation, the weighting factor of the self-weighting term is associated with the relative position of each pixel, particularly for those pixels surrounding the landmark point. Specifically, pixels that are closer to the landmark are assigned greater weights by $p_y$, while pixels that are farther away are assigned smaller weights by $p_y$, which advances the model in locating the position of the landmark.

The difference-modulating term is introduced for solving two problems: 1)
Inability to accurately evaluate the distribution of heatmap, as the value of ground truth ${p_y}$ is a probability. Existing loss functions, such as cross-entropy loss, cannot accurately evaluate the distribution of the heatmap.
2) Inability to distinguish easy/hard pixels.
Specifically, the difference-modulating term formulates the distance between the predicted value and the ground truth to guide the model towards converge, which has three contributions: 1) accurately evaluating the distribution of heatmap; 2)  able to distinguish easy/hard pixels;
3) solving the extreme imbalance between landmark pixels and back-ground pixels.
The experimental results show that our central loss significantly (p-value $< 0.05$) improves the performance of the landmark detection.

\subsection{Relationship with cross-entropy and focal loss}
Intuitively, when $r =0$, our central loss is equivalent to cross-entropy loss. During model training, the cross-entropy loss assigns equal weights to all pixels, which means that the loss function treats all misclassified samples equally.
A large number of misclassified negative samples may occupy most of the loss and dominate the gradient, causing landmark information to be ignored. Therefore, cross-entropy is difficult to solve the challenge of extreme imbalance between landmark pixels and background pixels. Our central loss can avoid the problem by balancing the
importance of positive/negative pixels using the novel difference-modulating term.
What's more, superior to focal loss, central loss can formulate the distance between the predicted value and the ground truth, which guides the model toward convergence. Additionally, central loss eliminates the need for an $\alpha$-balanced variant and allows for the adaptive adjustment of the proportion of various pixels based on their position information.

A more intuitive comparison is shown in Figure.~\ref{fig_lossCM}. For well-classified pixels, where the corresponding ground truth is 0.1 and 0.9, and the value of prediction is approximately equal to 0.1 and 0.9, the central loss can accurately evaluate and compute the loss (the loss value tends to 0). In contrast, even with perfect predictions, cross-entropy, and focal loss still provide a high loss for the model. These unconfident loss values can guide the model in the wrong direction, ultimately affecting its coverage and accuracy.

\noindent\textbf{ \textit{Discussion of the Innovation} }:

1) Central loss can accurately evaluate the distribution of heatmap, which addresses the challenge of inability to accurately evaluate the distribution of heatmap for heatmap-based landmark detection tasks.

2) Central loss can distinguish easy and hard pixels, which enables the model to pay more attention to the hard negatives and less attention to the easy samples, which addresses the challenge of inability to distinguish easy/hard pixel.

3) Central loss is associated with the relative position of each pixel, which addresses the challenge of extreme imbalance between landmark pixels and back-ground pixels by assigning greater weights to pixels that are closer to the landmark, while pixels that are farther away are assigned smaller weights.



\section{Coordinated Transformer}
\subsection{CoorTransformer}
While Transformer has achieved remarkable results in various domains~\cite{valanarasu2021medical,hatamizadeh2022unetr}, it still faces some challenges. For a medical image, the key elements are usually equal to image pixels ($H \times W$) and can be very large, which leads to the size of attention being quadratic large ($HW \times HW$), and further makes the convergence of attention tedious. Therefore, Transformer requires long training schedules before convergence is achieved, and the attention weights learned by Transformer are also almost uniformly distributed across all the pixels in the feature maps, especially for sparse landmarks. Due to the above mentioned problems, existing transformers are not efficient enough to extract meaningful information from the limited and sparse landmarks.

To overcome the above challenges, we introduce the coordination information into Transformer and propose a novel Coordinated Transformer (CoorTransformer). Specifically, as shown in Figure.~\ref{model}, CoorTransformer models the relations among landmarks effectively under the guidance of the location information in the feature maps. These focused locations are determined by sampling points which are learned from the queries by a coordinate module. Formally, given the input feature map $x \in {\mathbb{R}^{C \times H\times W}}$, the uniform grid of points $p \in {\mathbb{R}^{2 \times H\times W}}$ are computed as the coordination information through the coordinate
module. The values of coordinate points are linearly spaced 2D coordinates and are normalized to the range $[-1; +1]$, in which $(-1;-1)$ indicates the top-left corner and $(+1; +1)$ indicates the bottom-right corner. Afterward, coordinate-index features
 $x^p \in {\mathbb{R}^{C \times H\times W}}$ are sampled at each coordinate based on the coordinate information.

Different from the existing Transformer, the Query sequence  $Q^p \in \mathbb{R}^{c\times H\times W}$ and Key sequences  $K^p \in \mathbb{R}^{c\times H\times W}$ inside CoorTransformer are computed through coordinate-index features $x^p \in {\mathbb{R}^{C \times H\times W}}$,  By utilizing the input Query, Key, Value sequence $Q$, $K$ and $V$ . the  attention is computed as
\begin{equation}
    \text{Structure-aware Attention}(Q^p,K^p,V) = \text{softmax}(\frac{{Q^p}{K^p}^T}{\sqrt{d}})V,
\end{equation}
where $C$ is the number of channels, $H$ and $W$ represent the spatial dimension. And inside the attention layer, for each Query element, its similarity to all Key elements is calculated and then normalized via softmax, which is used to multiply the Value elements to achieve aggregated output and further establish long-range relations among each pixel.

Our CoorTransformer has three major advantages: 1) under the guidance of landmark coordination information, our attention can focus more on the sparse landmarks; 2) the computed attention is structure-aware, which can learn the underlying relationship between landmarks and incorporate rich structure knowledge. 3) With the information of coordinate index, the convergence of attention weights is faster, avoiding the defect that Transformer is difficult to converge in sparse
representation learning.


\begin{figure}[t]
  \centering  \includegraphics[width=0.5\textwidth]{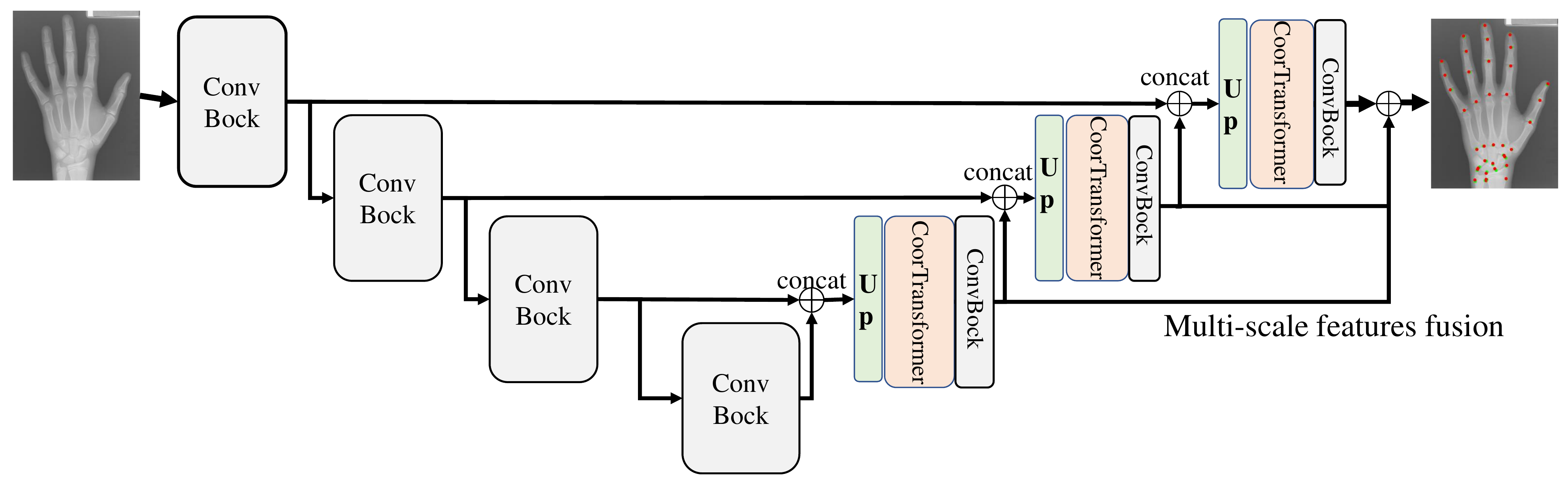}
  \caption{A graphical illustration of Coordinated Transformer based detector, which overcomes the drawbacks of neglecting global sensing of spatial structure present in existing methods.
}
  \label{fig3}
\end{figure}

\begin{table*}[]
\centering
  \caption{Quality metrics of different models on head, hand, and chest datasets show that CoorTransformer achieves the best performance. * represents the performances copied from the original paper. }
\makebox[1 \textwidth][c]{
\resizebox{1.0 \textwidth}{!}{
\begin{tabular}{c|ccccc|cccc|cccc}
\hline
\multirow{3}{*}{Method} & \multicolumn{5}{c|}{Head}                                                                                                          & \multicolumn{4}{c|}{Hand}                                                                              & \multicolumn{4}{c}{Chest}     \\ \cline{2-14}
                        & \multicolumn{1}{c|}{\multirow{2}{*}{MRE}} & \multicolumn{4}{c|}{SDR(\%)}                                                           & \multicolumn{1}{l|}{\multirow{2}{*}{MRE}} & \multicolumn{3}{c|}{SDR(\%)}                               & \multicolumn{1}{c|}{\multirow{2}{*}{MRE}} & \multicolumn{3}{c}{SDR(\%)}                              \\ \cline{3-6} \cline{8-10} \cline{12-14}
                        & \multicolumn{1}{c|}{}                     & \multicolumn{1}{c|}{2mm} & \multicolumn{1}{c|}{2.5mm} & \multicolumn{1}{c|}{3mm} & 4mm & \multicolumn{1}{c|}{}                     & \multicolumn{1}{c|}{2mm} & \multicolumn{1}{c|}{4mm} & 10mm & \multicolumn{1}{c|}{}                     & \multicolumn{1}{c|}{3px} & \multicolumn{1}{c|}{6px} & 9px \\ \hline
Ibragimov et al. \cite{ibragimov2014shape}*        & \multicolumn{1}{c|}{1.84}   & \multicolumn{1}{c|}{68.13}    & \multicolumn{1}{c|}{74.63}      & \multicolumn{1}{c|}{79.77}    &86.87     & \multicolumn{1}{c|}{--}                     & \multicolumn{1}{c|}{--}    & \multicolumn{1}{c|}{--}    &  --    & \multicolumn{1}{c|}{--}                     & \multicolumn{1}{c|}{--}    & \multicolumn{1}{c|}{--}    & --    \\ \hline
\v{S}tern et al. \cite{vstern2016local}*           & \multicolumn{1}{c|}{--}                   & \multicolumn{1}{c|}{--}    & \multicolumn{1}{c|}{--}      & \multicolumn{1}{c|}{--}    &--     & \multicolumn{1}{c|}{0.80}                     & \multicolumn{1}{c|}{92.20}    & \multicolumn{1}{c|}{98.45}    &99.83      & \multicolumn{1}{c|}{--}                     & \multicolumn{1}{c|}{--}    & \multicolumn{1}{c|}{--}    &--     \\ \hline
Lindner et al. \cite{lindner2014robust}*         & \multicolumn{1}{c|}{1.67}                 & \multicolumn{1}{c|}{70.65}    & \multicolumn{1}{c|}{76.93}      & \multicolumn{1}{c|}{82.17}    &\multicolumn{1}{c|}{89.85}     & \multicolumn{1}{c|}{0.85}                     & \multicolumn{1}{c|}{93.68}    & \multicolumn{1}{c|}{98.95}    &99.94      & \multicolumn{1}{c|}{--}                     & \multicolumn{1}{c|}{--}    & \multicolumn{1}{c|}{--}    &--     \\ \hline
Urschler et al. \cite{urschler2018integrating}*         & \multicolumn{1}{c|}{--}                   & \multicolumn{1}{c|}{70.21}    & \multicolumn{1}{c|}{76.95}      & \multicolumn{1}{c|}{82.08}    & 89.01    & \multicolumn{1}{c|}{0.80}                     & \multicolumn{1}{c|}{92.19}    & \multicolumn{1}{c|}{98.46}    & 99.95     & \multicolumn{1}{c|}{--}                     & \multicolumn{1}{c|}{--}    & \multicolumn{1}{c|}{--}    &--     \\ \hline
Payer et al. \cite{payer2019integrating}*          & \multicolumn{1}{c|}{--}                   & \multicolumn{1}{c|}{73.33}    & \multicolumn{1}{c|}{78.76}      & \multicolumn{1}{c|}{83.24}    &89.75     & \multicolumn{1}{c|}{\bf 0.66}                     & \multicolumn{1}{c|}{94.99}    & \multicolumn{1}{c|}{99.27}    & {\bf 99.99}      & \multicolumn{1}{c|}{--}                     & \multicolumn{1}{c|}{--}    & \multicolumn{1}{c|}{--}    &--     \\ \hline
Zhu et al. \cite{zhu2021you}*                  & \multicolumn{1}{c|}{1.54}                 & \multicolumn{1}{c|}{77.79}    & \multicolumn{1}{c|}{84.65}      & \multicolumn{1}{c|}{89.41}    & {94.93}     & \multicolumn{1}{c|}{0.84}                     & \multicolumn{1}{c|}{95.40}    & \multicolumn{1}{c|}{ 99.35}    &99.75      & \multicolumn{1}{c|}{5.57}                     & \multicolumn{1}{c|}{57.33}    & \multicolumn{1}{c|}{82.67}    & 89.33    \\ \hline
Our    & \multicolumn{1}{c|}{\bf 1.43}                 &  \multicolumn{1}{c|}{\bf 79.28}    & \multicolumn{1}{c|}{\bf 85.43}   & \multicolumn{1}{c|}{\bf 90.32}    &  {\bf 95.33}    & \multicolumn{1}{c|}{0.76}                     & \multicolumn{1}{c|}{ \bf 95.48}    & \multicolumn{1}{c|}{\bf 99.35}    & 99.86  & \multicolumn{1}{c|}{\bf 4.58}                     & \multicolumn{1}{c|}{\bf 63.82}    & \multicolumn{1}{c|}{\bf 89.84}    & {\bf 93.90}   \\ \hline
\end{tabular}
}
}
\label{tab:atlas}
\end{table*}
\subsection{CoorTransformer-based detector}
With the advanced CoorTransformer, we propose a CoorTransformer-based detector for landmark detection, which can inherently extract the landmark features and exploit the underlying relationship between landmarks for incorporating rich structural knowledge.
The architecture of the detector is shown in Figure.~\ref{fig3}, which consists of two stages and skip connections.
The encoder stage is constructed by a series of convolutional blocks, each convolutional block consists of a convolutional layer, batch normalization, and a Relu layer for feature extraction and down-sampling.
The decoder stage is constructed by a series of up-sampling layers and CoorTransformer modules with different numbers of attention heads. Meanwhile, spatial and channel attention is used in the convolution block of the decoder stage for feature fusion.
Behind the decoder stage, a classification head is attached. The classification head consists of a convolutional layer for getting the final detection result.

\noindent\textbf{ \textit{Discussion of the Innovation} }:

1) For the first time, coordinate information has been incorporated into the Transformer for landmark detection, enabling the exploitation of underlying relationships between sparse landmarks and the incorporation of rich structural knowledge.

2) CoorTransformer has an advantage in focusing more on sparse landmarks. With the guidance of coordinate information, CoorTransformer overcomes the challenge of Transformer struggling to converge in sparse representation learning.

3) CoorTransformer is structure-aware, allowing it to learn the underlying relationships between landmarks and incorporate rich structural knowledge.

4) Our detector, based on CoorTransformer, addresses the challenge of neglecting global spatial structure information in existing anatomical landmark detection methods.



\section{Experiments}
\subsection{Datasets and implementation details}
{\bf Cephalometric Landmark Dataset~\cite{wang2016benchmark}:}
The dataset consists of 400 cephalometric radiographs with 19 manually labeled landmarks by two doctors in each image, and the ground truth is the average of annotations of the two doctors. Each image is of size $1935 \times 2400$ with a resolution of $0.1mm \times 0.1mm$
We randomly choose 150 images for training and 250 images for testing. During training, we resize the original image to the size of $416 \times 512$ to keep the ratio of width and height.

{\bf Digital Hand Atlas Dataset~\cite{zhu2021you}:} The hand dataset is also a public dataset that contains 909 X-ray images of various sizes. And each image with 19 manually labeled landmarks.  During training, we resize images to the shape of 512 $\times$ 384. And 609 images are used for training and the other 300 images are used for testing.

{\bf Pulmonary Chest X-Ray Dataset ~\cite{jaeger2013automatic}:} This data is a public chest dataset that contains two subsets. We select the China set and exclude cases that are labeled as abnormal lungs(diseased lungs) to form our experimental dataset. Finally, the chest dataset has 279 X-ray images. Each image with six landmarks. During training, we randomly select 229 images used for training and the last 50 images for testing. And all of the input images are resized to the shape of 512$\times$512.


{\bf Training details:}
The number of CoorTransformer blocks is [1, 1, 1, 1], and the attention heads number is [8, 4, 2, 1]. The channel expansion factor is $\lambda  = 4$. The network is trained under the supervision of central loss (weighting parameter $r$ is set to 2). Our model is implemented using PyTorch \cite{paszke2019pytorch} and trained end-to-end with Adam \cite{kingma2014adam} optimization. In the training phase, the learning rate is initially set to 0.0001 and a cyclic scheduler is used to decrease the learning rate from 1e-3 to 1e-4 dynamically.
For evaluation, the inference model is chosen as the one with minimum validation loss.
The experiments were carried out on one NVIDIA RTX A4000 GPU with 16G memory and the batch size is set to 2. And mean radial error (MRE) that measures the Euclidean distance between prediction and ground truth and successful detection rate (SDR) with various settings are used as evaluation metrics.



\subsection{Comparison with state-of-the-art methods}
 CoorTransforme-based detector outperforms state-of-the-art landmark detection methods including Ibragimov et al.~\cite{ibragimov2014shape}, Stern et al.~\cite{vstern2016local}, Lindner et al.~\cite{lindner2014robust}, Urschler et al. \cite{urschler2018integrating}, Payer et al. \cite{payer2019integrating}, and Zhu et al. \cite{zhu2021you}.
The results of the  CoorTransformer and comparison methods are shown in Table~\ref{tab:atlas}.
On the head dataset, our CoorTransformer achieves the best accuracy (1.44 in MRE, and 79.28\%, 85.43\%, 90.32\%, 95.33\% in SDR with  2mm, 2.5mm, 3mm, 4mm thresholds). On the hand dataset, our CoorTransformer achieves the best accuracy (95.48\%, 99.35\%) in SDR with 2mm, 4mm threshold. On the hand dataset, our CoorTransformer achieves the best MRE accuracy (4.58) and the best SDR accuracy (63.82\%, 89.84\%, 93.90\%) with 3px,  6px, and 9px thresholds, respectively.
These improvements convincingly demonstrate the effectiveness of CoorTransformer in exploiting the underlying relationship among landmarks and incorporating rich structural knowledge.

\begin{figure}
  \centering
\includegraphics[width=0.4\textwidth]{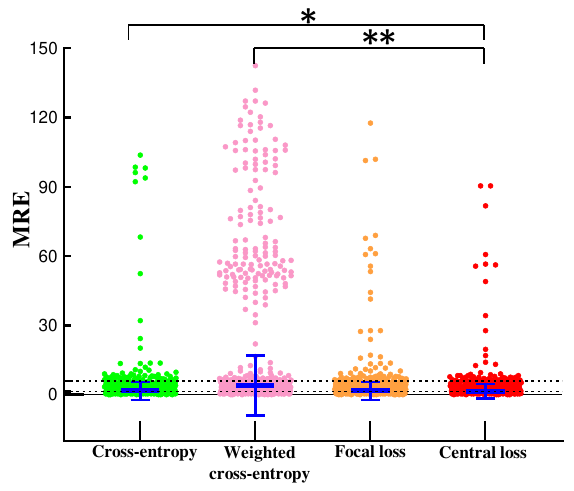}
  \caption{ Our central loss achieves better mean and standard deviation on the Cephalometric landmark dataset. The distribution of MRE of various Losses. Symbols ** and * represent significant differences between the corresponding method and our method, p $<$ 5e-5 and p $<$ 5e-2, respectively.}
  \label{fig_loss1}
\end{figure}

\begin{table}[]
\centering
  \caption{Quality metrics of different loss on head datasets. Experimental results demonstrate that central loss achieves the highest performance.}
\begin{tabular}{c|c|cccc}
\hline
\multirow{2}{*}{Method} & \multirow{2}{*}{MRE} & \multicolumn{4}{c}{SDR(\%)}                                                       \\ \cline{3-6}
                        &                      & \multicolumn{1}{c|}{2mm} & \multicolumn{1}{c|}{2.5mm} & \multicolumn{1}{c|}{3mm} & 4mm \\ \hline
Cross-entropy loss  &  1.78  & \multicolumn{1}{c|}{76.59}  & \multicolumn{1}{c|}{82.57}    & \multicolumn{1}{c|}{87.39}  & 92.91  \\ \hline
Weighted cross-entropy loss  &  4.02  & \multicolumn{1}{c|}{63.24}  & \multicolumn{1}{c|}{72.06}    & \multicolumn{1}{c|}{80.00}  & 89.09  \\ \hline
Focal loss  &  1.68  & \multicolumn{1}{c|}{78.21}  & \multicolumn{1}{c|}{83.75}    & \multicolumn{1}{c|}{87.96}  & 93.24  \\ \hline
Central loss  &  {\bf 1.59}  & \multicolumn{1}{c|}{\bf 79.60}  & \multicolumn{1}{c|}{\bf 85.37}    & \multicolumn{1}{c|}{\bf 89.60}  & {\bf 94.61}  \\ \hline
\end{tabular}
  \label{table_loss1}
\end{table}

\begin{figure*}[!t]
  \centering
\includegraphics[width=0.8\textwidth]{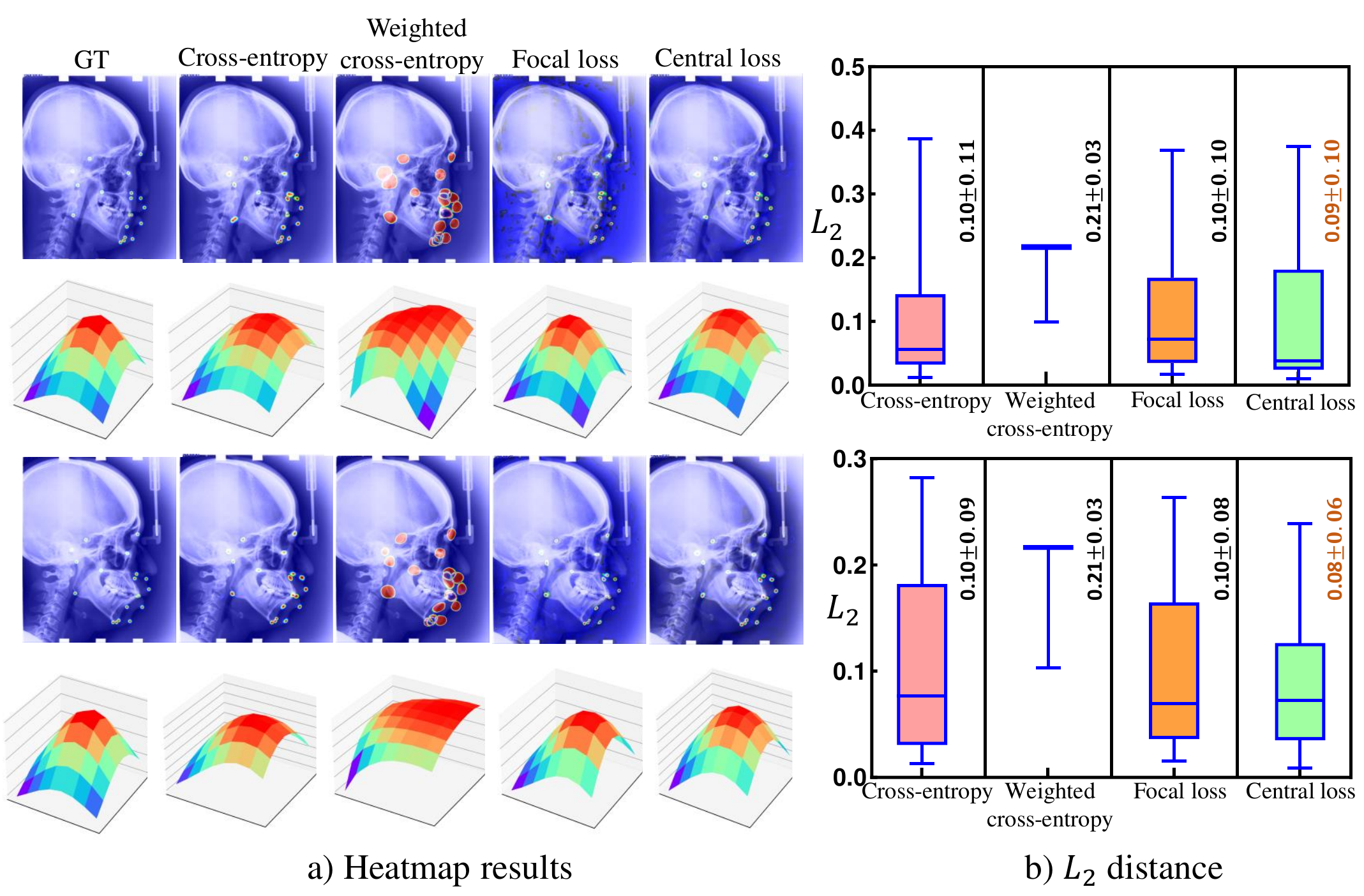}
  \caption{ Our novel central loss outperforms other losses and helps the baseline model achieve the highest accuracy.
  a) The predicated heatmap from various loss functions; b) The statistical analysis (mean and standard deviation) of $L_2$ distance.
  }
  \label{Ab}
\end{figure*}
\begin{figure}[t]
  \centering
    \includegraphics[width=0.4\textwidth]{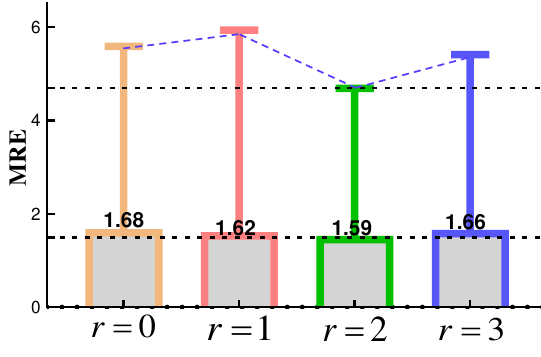}
  \caption{Statistical analysis (mean and standard deviation) of MRE for central loss across varying $r$ values.
}
  \label{fig_loss2}
\end{figure}

\subsection{Analysis of central loss}
To analyze the effectiveness of our central loss for solving the challenges associated with heatmap-based landmark detection, we conduct extensive studies on the Cephalometric Landmark dataset using representative loss functions including cross-entropy loss, weighted cross-entropy loss, and focal Loss. Meanwhile, to ensure an accurate and fair evaluation of each loss function, we removed other interfering components, such as CoorTransformer module, and exclusively employed Unet as the detection model.

\subsubsection{Effectiveness of central loss}
Table~\ref{table_loss1} and Figure.~\ref{fig_loss1} show that our central loss achieved the best quantitative results on the Cephalometric Landmark dataset. Table~\ref{table_loss1} lists the quantitative evaluation result of our central loss and other loss functions from a holistic perspective. Our central loss achieved the best MRE and SDR, which demonstrates that central loss is more computationally confident for the heatmap-based landmark detection task than other well-established loss functions. Furthermore, from the distribution of MRE of various losses in Figure.~\ref{fig_loss1}, our central loss also obtained a better mean and standard deviation, which further confirmed that central loss can effectively discount the effect of easy negatives, focusing all attention on the hard negative pixel under extreme imbalance conditions.
Meanwhile, as shown in Table~\ref{table_loss1}, the paired $t$-tests between central loss and the other three losses were performed on MRE value. The $p$-Values ($< 0.05$) of paired $t$-tests show that the difference between central loss and the other two base losses is significant, which also proves the superiority of central loss in landmark detection.

To gain a more comprehensive understanding of the central loss, an analysis was conducted on the distribution of landmarks. Figure~\ref{Ab} presents the predicted results obtained using different loss functions. It is evident that the weighted cross-entropy resulted in the poorest predictions with the highest error rate, even worse than that of the cross-entropy loss. This observation suggests that simply adding weights to landmarks is inadequate for addressing the challenges that arise in landmark detection.
In terms of quantitative evaluation of the alignment between predicted results and ground truth, the average $L_2$ distance between the distribution of predicted results and ground truth are computed. Central loss exhibited the most robust match with the distribution of true landmarks, surpassing the performance of both cross-entropy loss and focal loss, which demonstrates that cross-entropy and focal loss cannot accurately evaluate the distribution of landmark and the feedback information mislead the coverage of model.

\begin{table}[]
\centering
  \caption{Statistical analysis for central loss across varying $r$ values. Experimental results demonstrate that central loss achieves the highest performance when $r=2$.}
\begin{tabular}{c|c|cccc}
\hline
\multirow{2}{*}{Value} & \multirow{2}{*}{MRE} & \multicolumn{4}{c}{SDR(\%)}                                                       \\ \cline{3-6}
                        &                      & \multicolumn{1}{c|}{2mm} & \multicolumn{1}{c|}{2.5mm} & \multicolumn{1}{c|}{3mm} & 4mm \\ \hline
$r=0$  &  1.68  & \multicolumn{1}{c|}{76.67}  & \multicolumn{1}{c|}{82.84}    & \multicolumn{1}{c|}{88.38}  & 93.62  \\ \hline
$r=1$  &  1.62  & \multicolumn{1}{c|}{79.26}  & \multicolumn{1}{c|}{85.18}    & \multicolumn{1}{c|}{\bf 89.83}  & {\bf 94.88}  \\ \hline
$r=2$  &  {\bf 1.59}  & \multicolumn{1}{c|}{\bf 79.60}  & \multicolumn{1}{c|}{\bf 85.37}    & \multicolumn{1}{c|}{ 89.60}  & {94.61}  \\ \hline
$r=3$  &  {1.66}  & \multicolumn{1}{c|}{ 77.28}  & \multicolumn{1}{c|}{ 83.05}    & \multicolumn{1}{c|}{ 88.29}  & { 93.60}  \\ \hline
\end{tabular}
\label{table_loss2}
\end{table}

\begin{figure}[]
  \centering  \includegraphics[width=0.5\textwidth]{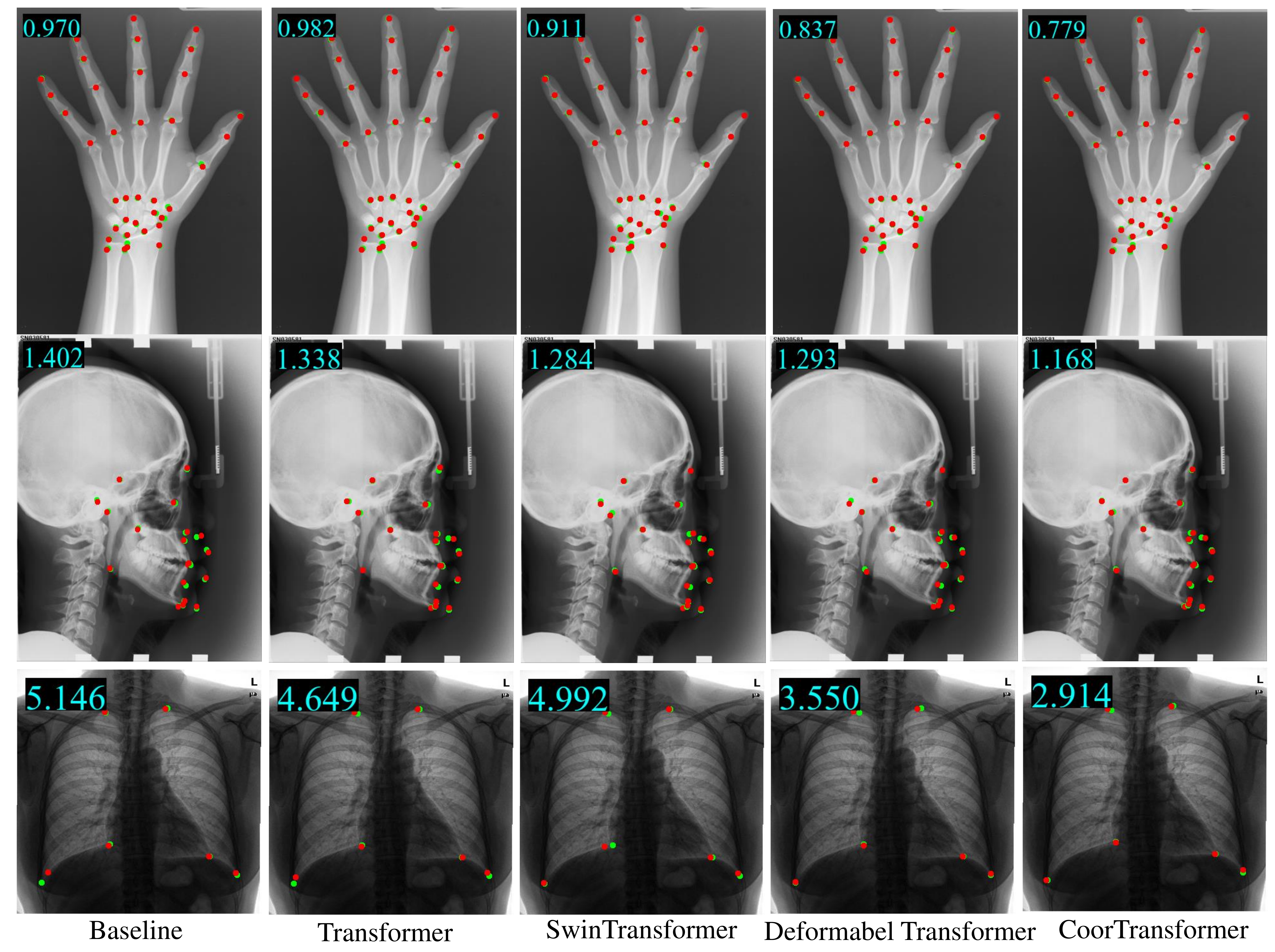}
  \caption{Experimental results demonstrate that our CoorTransformer outperforms other Transformers and helps the baseline model achieve the highest accuracy. The red points are the predicted landmarks while the green points are the ground truth labels.  On the top left corner of the image is the MRE value.}
  \label{seg}
\end{figure}
\begin{table*}[]
\centering
  \caption{Quality metrics of different Transformers on head, hand, and chest datasets. Experimental results demonstrate that CoorTransformer achieves the highest performance.}
\makebox[1 \textwidth][c]{
\resizebox{1.0 \textwidth}{!}{
\begin{tabular}{c|ccccc|cccc|cccc}
\hline
\multirow{3}{*}{Method} & \multicolumn{5}{c|}{Head}                                                                                                          & \multicolumn{4}{c|}{Hand}                                                                              & \multicolumn{4}{c}{Chest}     \\ \cline{2-14}
                        & \multicolumn{1}{c|}{\multirow{2}{*}{MRE}} & \multicolumn{4}{c|}{SDR(\%)}                                                           & \multicolumn{1}{l|}{\multirow{2}{*}{MRE}} & \multicolumn{3}{c|}{SDR(\%)}                               & \multicolumn{1}{c|}{\multirow{2}{*}{MRE}} & \multicolumn{3}{c}{SDR(\%)}                              \\ \cline{3-6} \cline{8-10} \cline{12-14}
                        & \multicolumn{1}{c|}{}                     & \multicolumn{1}{c|}{2mm} & \multicolumn{1}{c|}{2.5mm} & \multicolumn{1}{c|}{3mm} & 4mm & \multicolumn{1}{c|}{}                     & \multicolumn{1}{c|}{2mm} & \multicolumn{1}{c|}{4mm} & 10mm & \multicolumn{1}{c|}{}                     & \multicolumn{1}{c|}{3px} & \multicolumn{1}{c|}{6px} & 9px \\ \hline
Baseline
& \multicolumn{1}{c|}{1.78}
& \multicolumn{1}{c|}{76.59}
& \multicolumn{1}{c|}{82.57}
& \multicolumn{1}{c|}{87.39}
&92.91
& \multicolumn{1}{c|}{0.82}
& \multicolumn{1}{c|}{95.69}
& \multicolumn{1}{c|}{99.11}
& 99.64
& \multicolumn{1}{c|}{9.54}
& \multicolumn{1}{c|}{52.03}
& \multicolumn{1}{c|}{80.08}
& 91.46    \\ \hline
Transformer
& \multicolumn{1}{c|}{1.75}
& \multicolumn{1}{c|}{75.58}
& \multicolumn{1}{c|}{82.31}
& \multicolumn{1}{c|}{87.18}
&92.99
& \multicolumn{1}{c|}{0.84}
& \multicolumn{1}{c|}{95.96}
& \multicolumn{1}{c|}{99.27}
&99.60
& \multicolumn{1}{c|}{6.32}
& \multicolumn{1}{c|}{54.88}
& \multicolumn{1}{c|}{85.37}
& 91.87
\\ \hline
 Swin Transformer
& \multicolumn{1}{c|}{1.56}
& \multicolumn{1}{c|}{77.33}
& \multicolumn{1}{c|}{83.05}
& \multicolumn{1}{c|}{89.18}
& 95.14
& \multicolumn{1}{c|}{0.79}
& \multicolumn{1}{c|}{95.94}
& \multicolumn{1}{c|}{99.26}
& 99.66
& \multicolumn{1}{c|}{6.84}
& \multicolumn{1}{c|}{53.66}
& \multicolumn{1}{c|}{80.49}
& 90.24
\\ \hline
Deformable
Transformer
& \multicolumn{1}{c|}{1.55}
& \multicolumn{1}{c|}{76.32}
& \multicolumn{1}{c|}{83.18}
& \multicolumn{1}{c|}{88.69}
& 94.88
& \multicolumn{1}{c|}{0.82}
& \multicolumn{1}{c|}{96.03}
& \multicolumn{1}{c|}{99.14}
& 99.60
& \multicolumn{1}{c|}{5.55}
& \multicolumn{1}{c|}{50.00}
& \multicolumn{1}{c|}{82.21}
& 92.68
\\ \hline
 CoorTransformer
& \multicolumn{1}{c|}{\bf 1.43}
&  \multicolumn{1}{c|}{\bf 79.28}
& \multicolumn{1}{c|}{\bf 85.43}
& \multicolumn{1}{c|}{\bf 90.32}
&  {\bf 95.33}
& \multicolumn{1}{c|}{\bf 0.76}
& \multicolumn{1}{c|}{ \bf 95.48}
& \multicolumn{1}{c|}{\bf 99.35}
& {\bf 99.86}  & \multicolumn{1}{c|}{\bf 4.58}
& \multicolumn{1}{c|}{\bf 63.82}
& \multicolumn{1}{c|}{\bf 89.84}
& {\bf 93.90}
\\ \hline
\end{tabular}
}
}
\label{tab:Ab}
\end{table*}

\subsubsection{Impact of weighting parameter $r$}
Table.~\ref{table_loss2} and Figure.~\ref{fig_loss2} list the influence of $r$ on central loss. When $r=0$, the central loss is identical to the cross-entropy loss. In this case, Unet produces inferior results, indicating that cross-entropy loss is inadequate in addressing the issue of extreme sample imbalance and probability-based labels. However, according to the survey, cross-entropy loss is the most frequently used loss function in heatmap-based landmark detection tasks.
As the value of $r$ increases, the difference-modulating term formulates the distance between the predicted value and the ground truth to guide the model convergence. Additionally, this term can effectively distinguish the easy and hard pixels, making the model focus all attention on the hard negative examples for improving the performance of the model. Specifically, when $r=1$, Unet achieved improvements of  0.06, 2.59\%, 2.34\%, 1.45\%, and 1.26\% in MRE and SDR (2mm, 2.5mm, 3mm, and 4mm), respectively. When $r=2$, the improvements of Unet in MRE and SDR (2mm, 2.5mm, 3mm, and 4mm) are 0.09, 2.93\%, 2.53\%, 1.22\%, and 0.99\%, respectively. When $r=3$, the improvements of Unet in MRE and SDR (2mm, 2.5mm) are 0.02, 0.61\%, and 0.21\%, respectively.
Two innovations are responsible for the improved performance: 1) the difference-modulating term provides a reasonable and computationally confident solution to handling probability-based label values; 2) by combining the difference-modulating term with a self-weighting term, the model is able to focus more on hard negative examples and reduce attention on easier ones.


\subsection{Comparison with state-of-the-art Transformers}
We also compare CoorTransformer with state-of-the-art Transformers, including Transformer~\cite{vaswani2017attention}, Swin Transformer~\cite{liu2021Swin}, Deformable Attention Transformer~\cite{xia2022vision}, for demonstrating the superiority of the CoorTransformer in parse representations learning.

\subsubsection{Effective of CoorTransformer}
The results of the various Transformers are shown in Table~\ref{tab:Ab} and Figure.~\ref{seg}.
CoorTransformer beats all Transformers and helps the baseline (without Transformer module) model achieve the highest accuracy in head and hand datasets, which demonstrates that CoorTransformer is more effective in exploiting structural relationships between landmarks. Specifically, on the head data, the improvements in MRE and SDR (2mm, 2.5mm, 3mm and 4mm) are 0.35, 2.69\%, 2.86\%, 2.93\% and 2.42\%, respectively.
On the hand data, the improvements in MRE and SDR (2mm, 4mm, 10mm) are 0.06, 0.21\%, 0.24\%, and 0.22\%, respectively.
And on the chest data, the improvements in MRE and SDR (3px, 6px, 9px) are 4.87, 11.79\%, 9.76\%, and 2.44\%, respectively.
These improvements convincingly demonstrate the effectiveness of CoorTransformer in modeling content information among landmarks
and capturing structural information. Qualitative evaluation is also evaluated on three datasets. As it can be seen from Figure.~\ref{seg}, CoorTransformer also achieves a higher level of detection accuracy, which demonstrates that the innovations in our framework bring significant enhancement of Transformer in landmark detection.
\begin{figure}[]
  \centering  \includegraphics[width=0.4\textwidth]{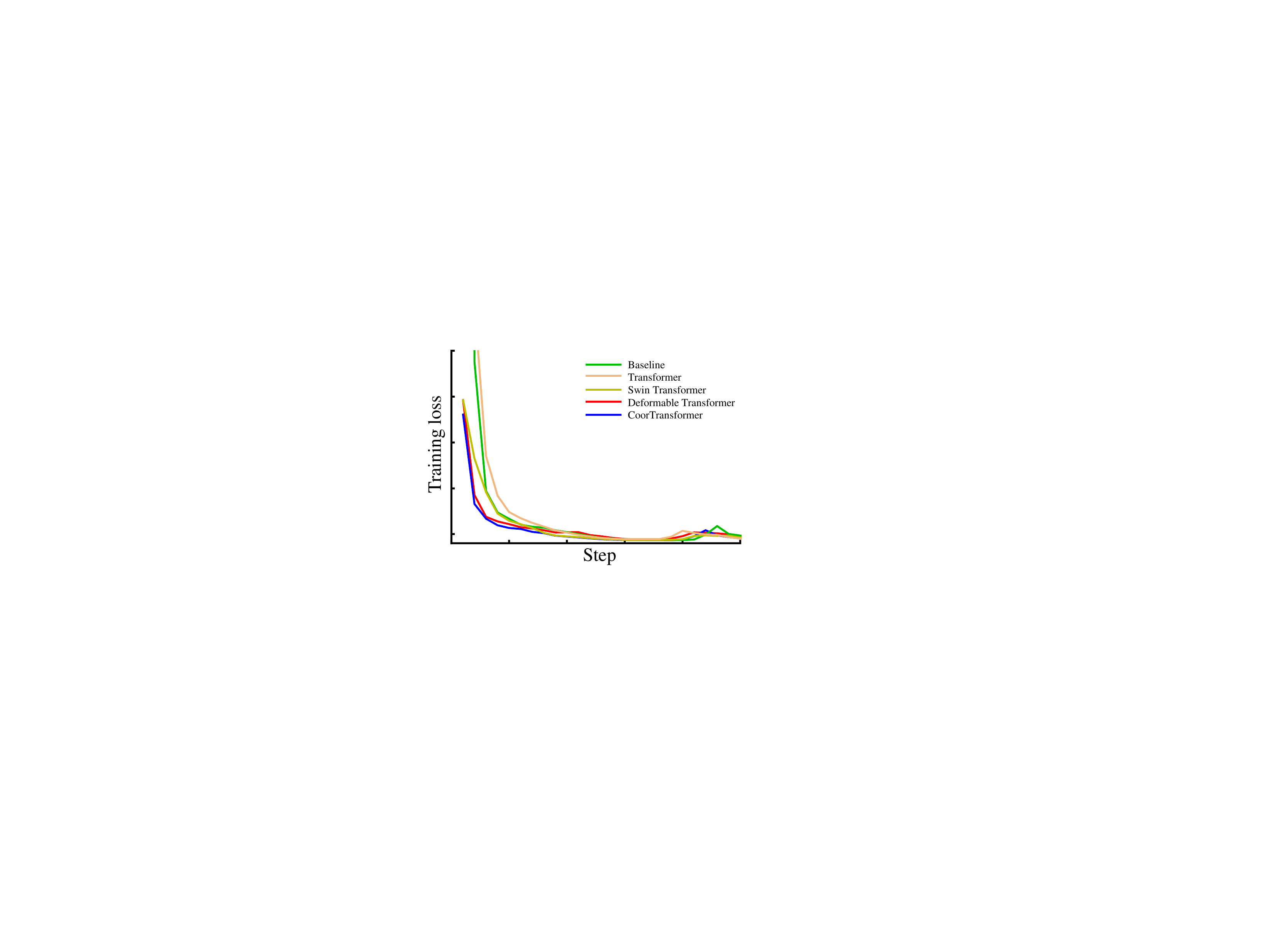}
  \caption{Experimental results demonstrate that CoorTransformer outperforms all other Transformers in terms of convergence.}
  \label{trainloss}
\end{figure}
\subsubsection{Advance of CoorTransformer} Figure.~\ref{trainloss} illustrates the loss convergence of several Transformers on the Head dataset. We can observe that the Transformer has the slowest convergence speed, even slower than the baseline. The major reason is that the vanilla self-attention module cannot effectively and reasonably model long-range dependencies among sparse landmarks. Our CoorTransformer outperforms all other Transformers in terms of convergence, which demonstrates that our CoorTransformer is more resilient and efficient when dealing with learning sparse representations.

\section{Conclusion}
For the first time, we have proposed a more effective and computational accurate central loss for the heatmap-based landmark detection, which addresses three challenges:
1) inability to accurately evaluate the distribution of heatmap; 2) inability to differentiate between easy and hard pixels; 3) difficulty dealing with extreme imbalance between landmark and background pixels. Furthermore, we have addressed the challenge of neglecting global sensing of spatial structure in existing landmark detection methods by designing a novel Coordinated Transformer. Our CoorTransformer incorporates coordination information when establishing long-range dependencies between pixels, making the attention more focused on sparse landmarks while incorporating the global spatial structure information. What's more,  CoorTransformer effectively avoids the defect that Transformer is difficult to converge in sparse representation learning. Experimental results from three landmark detection tasks demonstrate that our CoorTransformer can effectively extract the underlying relationship between landmarks for incorporating rich global structural knowledge. And our central loss significantly improves both the qualitative and quantitative performance of the landmark detection model.

\bibliographystyle{IEEEtran}
\bibliography{file}




\end{document}